# A Neural-Network Technique for Recognition of Filaments in Solar Images

V.V.Zharkova[1] and V.Schetinin[2]

[1]*Department of Cybernetics, University of Bradford, BD7 1DP, UK; e-mail: V.V.Zharkova@brad.ac.uk.*
[2]*Department of Computer Science, University of Exeter, EX4 4QF, UK; e-mail: V.Schetinin@ex.ac.uk*

**Abstract.** We describe a new neural-network technique developed for an automated recognition of solar filaments visible in the hydrogen H-alpha line full disk spectroheliograms. This technique allows neural networks learn from a few image fragments labelled manually to recognize the single filaments depicted on a local background. The trained network is able to recognize filaments depicted on the backgrounds with variations in brightness caused by atmospherics distortions. Despite the difference in backgrounds in our experiments the neural network has properly recognized filaments in the testing image fragments. Using a parabolic activation function we extend this technique to recognize multiple solar filaments which may appear in one fragment.

## 1. Introduction

Solar images observed from the ground and space-based observatories in various wavelengths were digitised and stored in different catalogues, which are to be unified under the grid technology. The robust techniques including limb fitting, removal of geometrical distortion, centre position and size standardisation and intensity normalisation were developed to put the $H_\alpha$ and Ca K lines full disk images taken at the Meudon Observatory (France) into a standardised form [1]. There is a growing interest to widespread ground-based daily observations of a full solar disk in the Hydrogen $H_\alpha$-line, which can provide important information on the long-term solar activity variations during months or years. The project European Grid of Solar Observations [2] was designed to deal with the automated detection of various features associated with solar activity, such as: sunspots, active regions and filaments, or solar prominences.

Filaments are the projections on a solar disk of prominences seen as very bright and large-scale features on the solar limb [1]. Their location and shape does not change very much for a long time and, hence, their lifetime is likely to be much longer then one solar rotation. However, there are visible changes seen in the filament elongation, position with respect to an active region and magnetic field configuration. For this reason the automated detection of solar filaments is a very important task to tackle in sense of understanding the physics of prominence formation, support and disruptions.

Quite a few techniques were explored for a different level of the feature detection such as: the rough detection with a mosaic threshold technique [3], the image segmentation and region growing techniques [4] – [7].

Artificial Neural Networks (ANNs) [8], [9] applied to the filament recognition problem, usually require a representative set of the image data available for training. The training data have to represent the image fragments, depicting filaments on different conditions, under which the ANN has to solve the recognition problem. For this reason the number of training examples must be large. However, the image fragments are still taken from the solar catalogues manually.

In this paper we describe a new neural-network technique, which is able to learn to recognize filaments from a few image fragments labelled manually. Despite the difference in backgrounds in selected fragments the trained network has properly recognized filaments in the testing image fragments. Using a parabolic activation function, below we extend this technique to recognize multiple solar filaments, which may appear in a single fragment.

## 2. A Recognition Problem

First, let us introduce the image data as a $n \times m$ matrix $\mathbf{X} = \{x_{ij}\}$, $i = 1, \ldots, n$, $j = 1, \ldots, m$, consisting of pixels whose brightness ranges between 1 and 255. This matrix depicts filament, which a dark elongated feature observable on the solar surface with a higher background brightness. Then, a given pixel $x_{ij} \in \mathbf{X}$ may belong to a filament region, class $\Omega_0$, or to a non-filament region, class $\Omega_1$.

Note, that the brightness of non-filament region varies significantly over the solar disk. We can make a realistic assumption that the influence of neighbouring pixels on the central pixel $x_{ij}$ is restricted to $k$ elements. Then we can easily define a triangular window, $k \times k$ matrix $\mathbf{P}^{(i,j)}$, with central pixel $x_{ij}$ and $k$ nearest neighbours. The background of the filament elements is assumed to be additive to $x_{ij}$ that allows us to evaluate and subtract it from the brightness values of all the elements of matrix $\mathbf{P}$.

Now we can define a background function $u = \varphi(\mathbf{X}; i, j)$, which reflects a total contribution of background elements to the pixel $x_{ij}$. Parameters of this function can be estimated from image data $\mathbf{X}$.

In order to learn background function $\varphi$ and then decide whether a pixel $x$ is a filament element or not, we can use image fragments whose pixels are manually labelled and assigned either to class $\Omega_0$ or class $\Omega_1$. A natural way to do this is to use a neural network which is able to learn recognising filaments from a few number of the labelled image fragments.

## 3. The Neural-Network Technique for Filament Recognition

In general, neural networks perform a threshold technique of pattern recognition and therefore this technique must take into account a variability of background elements in images. Correspondingly, the idea behind our method is to use additional information on the contribution of variable background elements, which is represented by the function $u = \varphi(\mathbf{X}; i, j)$. This function, as we assume, can be learnt from the image data.

One of the possible ways to estimate the function $\varphi$ is to approximate its values for each pixel $x_{ij}$ of a given image $\mathbf{X}$. For a filament recognition, we can use either a parabolic or linear approximation of this function. Whilst the first type is suitable for small size image fragments, the second one is used for relatively large fragments of the solar disk. Below we describe our neural-network technique exploiting the last type of approximation.

For image processing, the proposed algorithm exploits a standard sliding technique for which a given image matrix $\mathbf{X}$ is transformed into a column matrix $\mathbf{Z}$ consisting of $q = (n - k + 1)(m - k + 1)$ columns $\mathbf{z}^{(1)}, \ldots, \mathbf{z}^{(q)}$). Each column $\mathbf{z}$ presents the $r$ pixels taken from the matrix $\mathbf{P}$, where $r = k^2$. The pixels $x_{11}, x_{12}, \ldots, x_{1k}, \ldots, x_{k1}, x_{k2}, \ldots, x_{kk}$ of this matrix are placed in the columns of the matrix $\mathbf{Z}$ so that the central element of $\mathbf{P}$ is located in the $(r + 1)/2$ position of the column $\mathbf{z}$.

Let us now introduce a feed-forward ANN consisting of the two hidden and one output neurons as depicted in Figure 1. The first hidden neuron is fed by $r$ elements of column vector $\mathbf{z}^{(j)}$. The second hidden neuron evaluates the value $u_j$ of a background for the $j$th vector $\mathbf{z}^{(j)}$. The output neuron makes a decision, $y_i = \{0,1\}$, on the central pixel in the column vector $\mathbf{z}^{(j)}$.

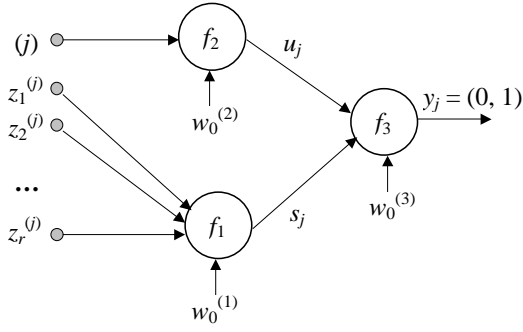

**Fig 1.** The feed-forward neural network consisting of the two hidden and one output neurons

Assuming that the first hidden neuron is fed by $r$ elements of the column vector $\mathbf{z}$, its output $s$ is calculated as follows:

$$s_j = f_1(w_0^{(1)}, \mathbf{w}^{(1)}; \mathbf{z}^{(j)}), j = 1, \ldots, q, \quad (1)$$

where $w_0^{(1)}$, $\mathbf{w}^{(1)}$, and $f_1$ are the bias term, weight vector and activation function of the neuron, respectively.

The activity of the second hidden neuron is proportional to the brightness of a background and can be described by the formula:

$$u_j = f_2(w_0^{(2)}, \mathbf{w}^{(2)}; j), j = 1, \ldots, q. \quad (2)$$

The bias term $w_0^{(2)}$ and weight vector $\mathbf{w}^{(2)}$ of this neuron are updated so that the output $u$ becomes an estimation of a background component contributed to the pixels of the $j$-th column $\mathbf{z}^{(j)}$. Parameters $w_0^{(2)}$ and $w^{(2)}$ may learn from the image data $\mathbf{Z}$.

Taking into account the outputs of the hidden neurons, the output neuron makes a final decision, $y_j \in (0, 1)$, for each column vector $\mathbf{z}^{(j)}$ as follows:

$$y_j = f_3(w_0^{(3)}, \mathbf{w}^{(3)}; s_j, u_j), j = 1, \ldots, q. \quad (3)$$

Depending on activities of the hidden neurons, the output neuron assigns a central pixel of the column $\mathbf{z}^{(j)}$ either to the class $\Omega_0$ or $\Omega_1$.

## 4. A Training Algorithm

In order to train the feed-forward ANN plotted in Figure 1, one can use the back-propagation algorithms, which provide a global solution. These algorithms require recalculating the output $s_j$ for all $q$ columns of matrix $\mathbf{Z}$ and for each the training epochs. However, there are some local solutions in which the hidden and output neurons are trained separately. Providing an acceptable accuracy, the local solutions can be found much easier than the global ones.

First, we need to fit the weight vector of the second hidden, or a "background" neuron that evaluates a contribution of the background elements. Figure 2 depicts an example, where the top left plot shows the image matrix $\mathbf{X}$ presenting a filament on the unknown background and the top right plot reveals the detected filament elements depicted in black. The two bottom plots in Figure 2 show the outputs $s$ (the left plot) and the weighted sum of $s$ and $u$ (the right one) plotted versus the columns of matrix $\mathbf{Z}$.

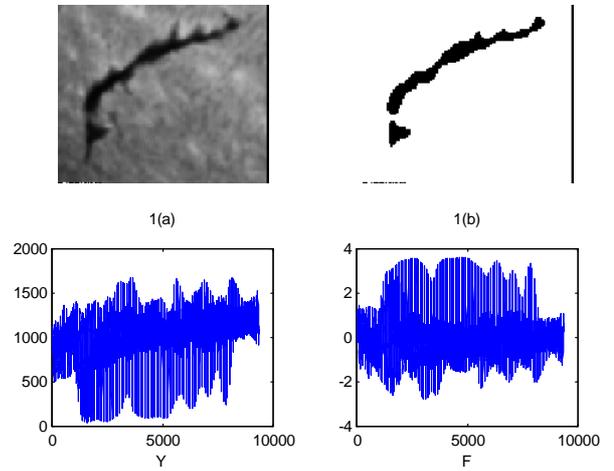

**Fig 2:** The example of image matrix $\mathbf{X}$ depicting the filament on the unknown background.

From the left top plot we see that the brightness of a background varies from the lowest level at the left bottom corner to the highest at the right top corner. Such variations of the background increase the output value $u$ calculated over the $q$ columns of matrix **Z**, see the increasing curve depicted at the bottom left plot. This plot shows us that the background component is changed over $j = 1, \ldots, q$ and can be fitted by a parabola..

Based on this finding we can define a parabolic activation function of the "background" neuron as:

$$u_j = f_2(w_0^{(2)}, \mathbf{w}^{(2)}; j) = \\ = w_0^{(2)} + w_1^{(2)} j + w_2^{(2)} j^2. \quad (4)$$

The weight coefficients $w_0^{(2)}$, $w_1^{(2)}$, and $w_2^{(2)}$ of this neuron can be fitted to the image data **Z** so that a squared error $e$ between the outputs $u_i$ and $s_i$ became minimal:

$$e = \Sigma_j (u_j - s_j)^2 = \\ = \Sigma_i (w_0^{(2)} + w_1^{(2)} j + w_2^{(2)} j^2 - s_j)^2 \\ \rightarrow min, j = 1, \ldots, q. \quad (5)$$

The desirable weight coefficients can be found with the least square method. Using a recursive learning algorithm [10], we can improve the evaluations of these coefficients due to robustness to non-Gauss noise in image data. So the "background" neuron can be trained to evaluate the background component $u$.

In the right bottom plot of Fig 2 we present the normalized values of $s_i$, which are no longer affected by the background component. The recognized filament elements are shown at the right top plot in the Fig 2. By comparing the left and right top plots in Fig 2 we can reveal that the second hidden neuron has successfully learnt to evaluate a background component from the given image data **Z**.

Before training the output neuron, the weights of the first hidden neuron are to be found. For this neuron a local solution is achieved for a set of the coefficients to be equal to 1. In case, if it would be necessarily to improve the recognition accuracy, one can update these weights by using the back-propagation algorithm.

After defining the weights for both hidden neurons, it is possible to train the output neuron, which makes the decisions between 0 or 1. Let us re-write the output $y_i$ of this neuron as follows:

$$y_j = 0, \text{ if } w_1 s_j + w_2 u_j < w_0, \text{ and} \\ y_j = 1, \text{ if } w_1 s_j + w_2 u_j \geq w_0. \quad (6)$$

Then the weight coefficients $w_0$, $w_1$, and $w_2$ can be fit in such way that the recognition error $e$ is minimal:

$$e = \Sigma_i |y_i - t_i| \rightarrow min, i = 1, \ldots, h, \quad (7)$$

where $|\cdot|$ means a modulus operator, $t_j \in (0, 1)$ is the $i$th element of a target vector **t** and $h$ is the number of its components, namely, the training examples.

In order to minimize the error $e$ one can apply any supervised learning methods, for example, the perceptron learning rule [8], [9].

## 5. Results and Discussion

The neural network technique described above was applied for recognition of dark filaments in solar images. The full disk solar images obtained on the Meudon Observatory (France) during the period of March-April 2002 were considered for the identification [1].

The fragments with filaments were picked from the Meudon full disk images taken for various dates and regions on the solar disk with different brightness and inhomogeneity in the solar atmosphere. There were 55 fragments selected depicting the filaments on a various background, one of them was used for training the ANN, the remaining 54 ones were used for testing the trained ANN. Visually comparing the resultant and origin images we can conclude that our algorithm recognised these testing filaments well.

Using a parabolic approximation we can now recognize the large image fragments which may contain multiple filaments as depicted at the left plot in Figure 3.

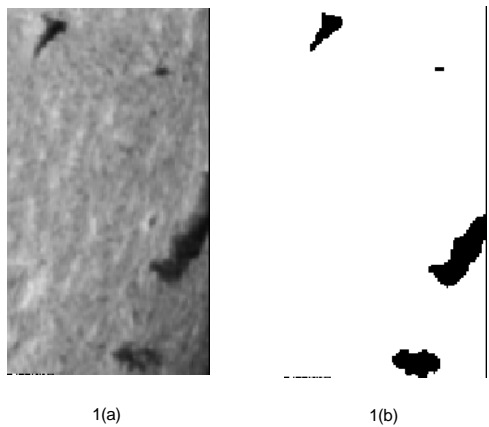

**Fig 3:** The recognition of the multiple filaments

The recognized filament elements depicted here in black are shown at the right plot. A visual comparison of the resulting and original images confirms that the proposed algorithm has recognized all four filaments being close to their location and shape.

## 6. Conclusions

The automated recognition of filaments on the solar disk images is still a difficult problem because of a variable background and inhomogeneities in the solar atmosphere. The proposed neural network technique can learn the recognition rules from a few images depicting the solar filament fragmented and labelled visually. The recognition rule has been successfully tested on the 54 other image fragments depicting filaments on different backgrounds. Despite the background differences, the trained neural network has properly recognized as single as multiple filaments presented in the testing image fragments Therefore, the proposed neural network technique can be effectively used for an automated recognition of filaments in solar images.